\documentclass[twocolumn]{NobArticle}

\usepackage{float}
\usepackage{placeins}
\usepackage{array}
\usepackage{colortbl}
\usepackage{tikz}

\newcommand{\ours}{D4GS}

\runninghead{D4GS: Efficient Gaussian Splatting SLAM with Depth Priors}
\footertext{Preprint}

\title{D4GS: Efficient Gaussian Splatting SLAM with Depth Priors}

\author{
    Da Li\textsuperscript{1}, 
    Yihan Zang\textsuperscript{1},
    Ivan Viola\textsuperscript{1},
    Zeyu Wang\textsuperscript{2}
    and Dominik Engel\textsuperscript{1}
}

\date{
    \textsuperscript{\textbf{1}}
    King Abdullah University of Science and Technology\\ \textsuperscript{\textbf{2}}
    The Hong Kong University of Science and Technology (Guangzhou) \\ 
    \textsuperscript{1}\texttt{\{da.li, yihan.zang, ivan.viola, dominik.engel\}@kaust.edu.sa}\\
    \textsuperscript{2}\texttt{zeyuwang@ust.hk }
}

\renewcommand{\maketitlehookd}{%
\begin{abstract}
    \noindent Achieving high-quality reconstruction and responsive camera tracking remains challenging for monocular RGB Gaussian splatting SLAM under a limited optimization budget. We present an online system that addresses this challenge by combining learned geometric priors with explicit geometric camera tracking, which efficiently performs camera tracking while maintaining the map to keep the predicted depth scale consistent. A frozen depth model serves as a virtual RGB-D sensor, providing depth estimation from monocular video. An explicit geometric solver then estimates camera pose through feature reprojection, and a streaming Gaussian mapper integrates the resulting observations into a dense scene representation. Extensive experiments on indoor and outdoor datasets show that our method significantly improves reconstruction quality over existing approaches, achieving over 10 FPS on a single RTX 4090 and delivering almost twice the throughput of the evaluated fastest prior SLAM systems on the 7-Scenes dataset.
\end{abstract}
}

\begin{document}

\small
\maketitle

\section{Introduction}
\label{sec:intro}

Monocular Gaussian SLAM aims to estimate the camera pose and build a
renderable scene map online using only an RGB image stream.
This capability supports immersive interaction, live scene capture,
and remote interaction in AR/VR and robotics.
3D Gaussian splatting (GS)~\citep{kerbl2023gs} provides an explicit
scene representation with efficient rendering and differentiable mapping,
making it a useful representation for dense
SLAM.
However, unlike RGB-D Gaussian SLAM methods~\citep{keetha2024splatam, yan2024gsslam, yugay2023gaussianslam, peng2024rtgslam, ha2024gsicp, zhu2025loopsplat}, without measured depth, the monocular GS SLAM system must jointly recover camera
pose, scene geometry, and the appearance, within a limited computational budget.

Methods that rely on sparse feature points or incremental map
optimization struggle to achieve high-fidelity mapping~\citep{matsuki2024monogs,huang2024photoslam}.
Learned dense priors help fill this gap, but integrating them into a consistent map introduces further challenges. Dense depth prediction and camera joint optimization add online computational cost~\citep{homeyer2024droidsplat,cheng2025s3po, zhang2025hi}. Depth determines where Gaussians are initialized, while camera poses determine whether observations from different frames align. Poor initialization requires additional map refinement, and inaccurate poses introduce conflicting observations that can blur the reconstructed scene. The challenge is therefore to obtain consistent dense depth and camera poses together within a limited online computational budget.

\begin{figure}[t]
  \centering
  \includegraphics[width=\linewidth]{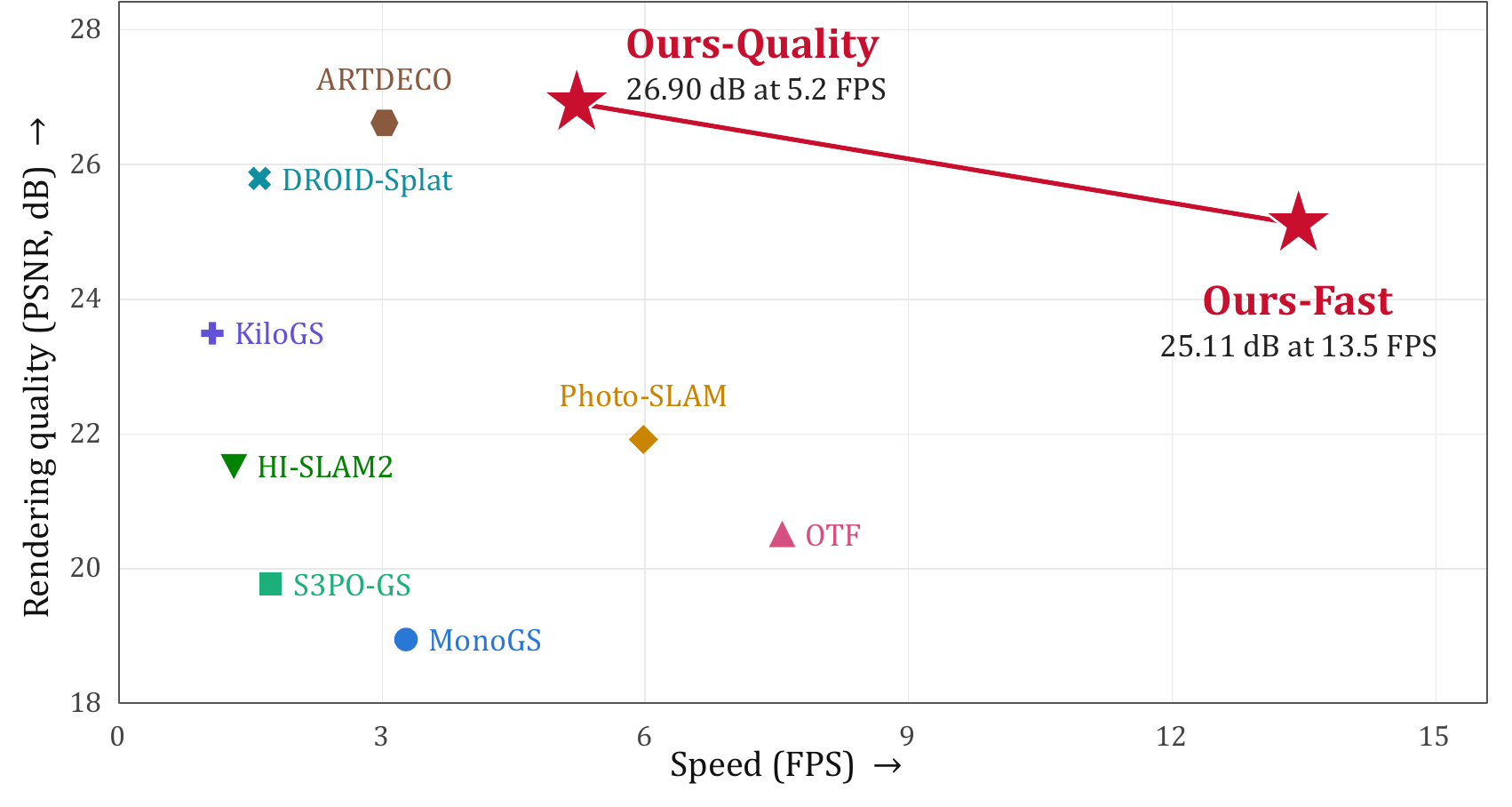}
   
  \caption{Quality--speed trade-off on 7-Scenes.
  Our method offers a better trade-off between rendering quality and speed
  than prior methods: \textbf{Ours-Quality} achieves the best rendering
  quality, while \textbf{Ours-Fast} is the fastest method at comparable
  quality.}
  \label{fig:teaser}
\end{figure}

Streaming 3D foundation models offer a promising starting point, predicting dense depth and camera poses as images arrive by maintaining temporal context~\citep{su2026xstreamvggt,zhuo2026streamvggt,
chen2026ttt3r,chen2026lingbotmap}. Their depth estimates provide dense surface information for Gaussian initialization, but we observe that errors in their accompanying poses can still cause cross-view misalignment. Single-frame depth models~\citep{hu2024metric3d,depth_anything_v2,depthanything3,piccinelli2025unidepthv2} provide an efficient alternative for dense geometry estimation without temporal context. However, their predicted scale can vary across frames, leading to inconsistent Gaussian initialization and depth supervision. Using these predictions for online mapping therefore requires reliable consistent depth scale alignment.

To address these challenges, we introduce D4GS, an efficient
online Gaussian SLAM framework that combines single-frame depth
priors with sparse feature-based geometric optimization.
Our scale anchoring aligns predicted depth with the existing
sparse map and suppresses frame-to-frame scale fluctuations
through temporal filtering.
The corrected depth jointly constrains camera pose estimation
with image feature observations and supports Gaussian
initialization and optimization.
Local bundle adjustment and loop closure maintain geometric
consistency among historical keyframes, while pose
synchronization propagates these corrections to Gaussian mapping.

Our system requires no additional training. We evaluate a fast configuration with 30 mapping iterations per keyframe and a quality configuration with 150. Our main contributions are:
\begin{itemize}
    \item \textbf{Online scale anchoring for monocular depth priors.}
    We introduce a scale anchoring method that aligns predicted
    depth with the evolving sparse map and regularizes the
    correction through temporal filtering.
    This mitigates frame-to-frame scale fluctuations and provides
    a shared depth prior for camera tracking and Gaussian mapping.

    \item \textbf{An efficient depth-assisted monocular Gaussian
    SLAM system.}
    We develop D4GS, integrating depth-constrained feature-based
    pose estimation with incremental Gaussian mapping. Ablations with multiple depth networks demonstrate compatibility with different depth priors.

    \item \textbf{Comprehensive evaluation of efficiency and
    rendering quality.}
    We evaluate rendering quality, trajectory accuracy, reconstruction geometry, and throughput on indoor and outdoor scenes. The two configurations expose the trade-off between mapping cost and rendering quality within the same system, as illustrated in Fig.~\ref{fig:teaser}.
\end{itemize}

\section{Related Work}
\label{sec:related_work}

\paragraph{Visual tracking without depth prior.}
Prior Gaussian SLAM methods differ in how they recover dense geometry and how they use it for camera tracking. These choices determine the balance between reconstruction quality and online computational cost.
MonoGS~\citep{matsuki2024monogs} estimates poses by photometric
alignment with the Gaussian map, making tracking dependent on
the quality of the evolving reconstruction.
Photo-SLAM~\citep{huang2024photoslam} combines
ORB-SLAM3~\citep{campos2021orbslam3} with Gaussian mapping.
CaRtGS~\citep{feng2025cartgs}, RP-SLAM~\citep{bai2024rpslam},
and SEGS-SLAM~\citep{wen2025segs} further develop this architecture
through improvements to computation allocation, map updates,
and scene modeling.
MGSO~\citep{hu2024mgso} instead initializes Gaussians from geometry
recovered by direct photometric odometry.
Without learned dense geometric priors, surface recovery relies
on multi-view image observations and parallax.
Regions with weak visual constraints may remain insufficiently
reconstructed at initialization, requiring interpolation or
further geometric refinement during mapping.

\paragraph{Learned geometry for tracking and mapping.}
Learned depth and pointmap predictions provide richer surface
information for tracking and mapping.
Splat-SLAM~\citep{sandstrom2024splatslam},
DROID-Splat~\citep{homeyer2024droidsplat}, and
HI-SLAM2~\citep{zhang2025hi} integrate learned dense depth and trajectory refinement.
MGS-SLAM~\citep{zhu2024mgsslam} uses multi-view depth supervision,
while SplatMAP~\citep{hu2025splatmap} updates Gaussian positions
using refined poses and depths from dense SLAM, and uses
reliability masks to prune or add Gaussians.
Some methods strengthen geometric consistency through
map-to-tracker feedback or optical flow constraints
~\citep{zhu2026mygosplat,seo2026gaussianflow,zhao2025pseudo},
and suppress unreliable observations in dynamic scenes
~\citep{zheng2025wildgs,li2025dy3dgs}.
These designs improve geometric estimation, but dense tracking and iterative refinement add computation to the online pipeline, limiting their application.

Feed-forward prediction offers a way to reduce this burden.
StreamGS~\citep{li2025streamgs} and Flash-Mono~\citep{zhang2026flash} predict
poses and Gaussian attributes before limited map refinement, but require intensive model retraining.
Pointmap and multi-view foundation models also support
tracking and mapping in OpenGS-SLAM~\citep{yu2025opengsslam}, ARTDECO~\citep{li2025artdeco}
and S3PO-GS~\citep{cheng2025s3po}, but with a high tracking cost.
GeoGS-SLAM~\citep{gao2026geogs},
VGGT-GS SLAM~\citep{han2026vggtgsslam}, and
M$^3$~\citep{ren2026m3} combine learned predictions with
geometric or photometric optimization to refine poses
and scene estimates by processing frames in batches.
OTF~\citep{meuleman2025onthefly} explores efficient incremental
Gaussian construction, but loses consistent geometry reconstruction.

Combining predicted geometry with explicit pose estimation
is particularly relevant to our work.
GigaSLAM~\citep{deng2025gigaslam} and
KiloGS-SLAM~\citep{yu2026robust} use predicted depth with
geometric solvers, while KiloGS-SLAM additionally
activates learned pixel matching when needed, but they mainly focus on large-scale scene reconstruction.
MonoEM-GS~\citep{kruzhkov2026monoemgs} uses iterative closest point (ICP) to align
predicted pointmap with the Gaussian representation.
These approaches demonstrate the value of predicted geometry for explicit camera pose estimation. D4GS focuses on the temporal scale consistency of single-frame depth predictions. It aligns these predictions with a sparse map and filters the scale correction, allowing the corrected depth to support both feature-based tracking and incremental Gaussian mapping efficiently.

\section{Method}
\label{sec:method}

\begin{figure*}[t]
  \centering
  \definecolor{d4scale}{HTML}{D9822B}\definecolor{d4scalebg}{HTML}{FBE3C8}
  \definecolor{d4track}{HTML}{3F6FA8}\definecolor{d4trackbg}{HTML}{DCE5F1}
  \definecolor{d4map}{HTML}{3E8F58}\definecolor{d4mapbg}{HTML}{D3E9D6}
  \newcommand{\figeq}[2]{{\hypersetup{linkcolor=#1}Eq.~(\ref{#2})}}% tag text: same colour as the tag, still a link
  \newcommand{\figeqs}[3]{{\hypersetup{linkcolor=#1}Eqs.~(\ref{#2},\,\ref{#3})}}
  \begin{tikzpicture}
    \node[anchor=south west, inner sep=0] (img) {\includegraphics[width=\linewidth]{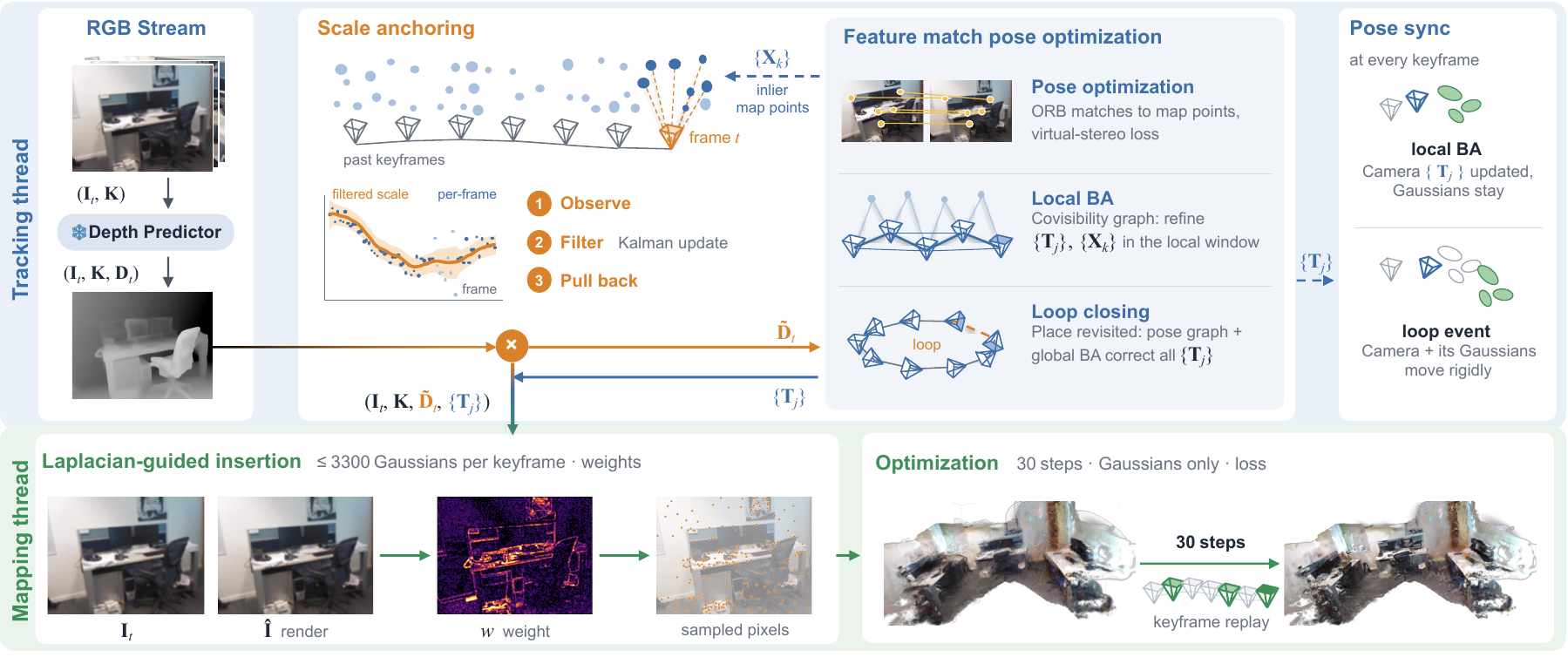}};
    \begin{scope}[x={(img.south east)}, y={(img.north west)},
        eqtag/.style={anchor=west, font=\sffamily\bfseries\fontsize{4.9}{5.4}\selectfont, inner xsep=2pt, inner ysep=1.3pt, rounded corners=2pt}]
      \node[eqtag, fill=d4scalebg, text=d4scale] at (0.4703,0.6290) {\figeq{d4scale}{eq:scale_filter}};
      \node[eqtag, fill=d4trackbg, text=d4track] at (0.7455,0.7949) {\figeq{d4track}{eq:poseba}};
      \node[eqtag, fill=d4mapbg, text=d4map] at (0.4151,0.2957) {\figeq{d4map}{eq:laplacian_insertion}};
      \node[eqtag, fill=d4mapbg, text=d4map] at (0.8139,0.2936) {\figeqs{d4map}{eq:mapping_loss}{eq:inverse_depth_loss}};
    \end{scope}
  \end{tikzpicture}
  \caption{Overview of our online Gaussian splatting SLAM system. (a) A frozen depth model predicts depth from the initial RGB frames. (b) Predicted depth aligned by scale anchoring, and image features support geometric camera tracking through correspondence matching and reprojection error minimization. (c) The Gaussian mapper integrates RGB frames, predicted depth, and estimated poses through keyframe selection after pose synchronization, Laplacian-guided Gaussian insertion, and 30-step refinement using RGB and depth losses with keyframe replay.}
  \label{fig:overview}
\end{figure*}

\ours{} processes a monocular video frame by frame using known intrinsics $\mathbf{K}$
(Fig.~\ref{fig:overview}).  For each frame $\mathbf{I}_t$, a single-frame depth network
first predicts a depth map $\mathbf{D}_t$.  
To reduce frame-to-frame scale variation, scale anchoring (Sec.~\ref{sec:method:scale}) aligns this prediction with the sparse map and temporally filters the correction. The tracker (Sec.~\ref{sec:method:track}) uses the RGB frame and corrected depth to estimate camera pose and maintain the sparse map through local bundle adjustment and loop closure. Before refining the Gaussian map, the mapper synchronizes its stored keyframe poses with the tracker (Sec.~\ref{sec:method:mapper}).

\subsection{Scene Representation}
\label{sec:method:representation}

D4GS maintains a sparse geometric map $\boldsymbol{\mathcal{M}}$ for camera
tracking and depth-scale estimation, together with a Gaussian map
$\boldsymbol{\mathcal{G}}$ for dense reconstruction and rendering. Both
representations are expressed in a shared world coordinate system.
We parameterize the camera pose at time $t$ by a world-to-camera
transformation
$\mathbf{T}_t = (\mathbf{R}^{\mathrm{cam}}_t,\mathbf{t}_t) \in SE(3)$.

\textbf{Sparse geometric map} consists of landmarks $\{\mathbf{X}_k \in\mathbb{R}^3\} $,
their associated feature descriptors, and observations in
tracking keyframes. We denote the observation set by
$
\boldsymbol{\mathcal{O}}
=
\{(j,k,\mathbf{u}_{jk})\},
$
where $\mathbf{u}_{jk}$ is the pixel location of landmark
$\mathbf{X}_k$ in keyframe $j$. These associations provide the
geometric constraints for pose estimation and bundle adjustment.
They also connect predicted depth to the existing reconstruction:
a matched landmark supplies a map-based depth estimate at its
corresponding pixel location, from which the depth-scale module
constructs its observations. \textbf{Gaussian map} is represented by a set of 3D Gaussians,
$\boldsymbol{\mathcal{G}}=\{\mathbf{g}_i\}_{i=1}^{N_G}$, with
$\mathbf{g}_i=(\boldsymbol{\mu}_i,\mathbf{q}_i,\mathbf{s}_i,
\alpha_i,\mathbf{c}_i)$,
where $\boldsymbol{\mu}_i\in\mathbb{R}^3$ is the mean,
$\mathbf{q}_i\in\mathbb{R}^4$ is the rotation quaternion,
$\mathbf{s}_i\in\mathbb{R}_{>0}^3$ contains the scales
along the three principal axes,
$\alpha_i\in[0,1]$ is the opacity, and
$\mathbf{c}_i\in\mathbb{R}^3$ is the color.

\subsection{Scale Anchoring}
\label{sec:method:scale}

To reduce frame-to-frame scale variations in predicted depth,
we estimate a correction relative to the sparse map and
regularize it over time using a scalar Kalman filter~\citep{kalman1960new}. At each frame, the filter requires two quantities: an observation
$m_t$ of the current scale discrepancy between the sparse map and
the predicted depth, and a variance $\nu_t$ indicating how
reliable this observation is.

Let $\mathbf{T}_{t-2}$ and $\mathbf{T}_{t-1}$ denote the
estimated poses of the two most recent consecutive frames.
Their relative motion is
$\mathbf{V}_{t-1}=\mathbf{T}_{t-1}\mathbf{T}_{t-2}^{-1}$.
We initialize the current world-to-camera pose with the constant-velocity model as $\bar{\mathbf{T}}_t=\mathbf{V}_{t-1}\mathbf{T}_{t-1}$.
Previous inlier landmarks are projected into the current view
using $\bar{\mathbf{T}}_t$  and matched to nearby ORB features. For each matched landmark with projected camera-space depth
$\bar{z}_{tk}$ at pixel $\mathbf{u}_{tk}$, we compute the
log-scale ratio
$r_{tk}=\log(\bar{z}_{tk}/\mathbf{D}_t(\mathbf{u}_{tk}))$.
The scale observation $m_t$ is the median of $\{r_{tk}\}_k$
weighted by descriptor similarity, and its variance $\nu_t$
is the squared robust standard deviation derived from the
Median Absolute Deviation (MAD) of $\{r_{tk}\}_k$, with details provided in the appendix.

We model the log-scale correction $\ell_t$ as a random walk
with process variance $q^2$ and maintain its uncertainty $P_t$:
\begingroup
\begin{equation}
\begin{gathered}
P_t^- = P_{t-1}+q^2,
\qquad\qquad
P_t = (1-\kappa_t)P_t^-, \\[2pt]
\kappa_t = \frac{P_t^-}{P_t^-+\nu_t},
\quad
\ell_t = \ell_{t-1}+\kappa_t(m_t-\ell_{t-1}),
\end{gathered}
\label{eq:scale_filter}
\end{equation}
\endgroup
where $P_t^-$ is the predicted variance and $\kappa_t$ is the
Kalman gain.
The process noise standard deviation $q$ specifies how much the
true log-scale correction is expected to change between two
consecutive frames.
We set $q=0.005$, i.e., about $0.5\%$ per frame, which favors
temporally smooth estimates while still tracking gradual scale
changes.

Because newly initialized landmarks inherit the corrected
depth, alignment to the evolving sparse map can accumulate
scale drift.
We use the predicted depth's approximately metric output as a soft
reference by attenuating large log-scale corrections with
$\gamma_t=\sigma^2/(\sigma^2+\ell_{t-1}^2)$,
where $\sigma=0.15$ controls the attenuation.
Both tracking and mapping receive the same corrected depth,
$\tilde{\mathbf{D}}_t=e^{\gamma_t\ell_t}\mathbf{D}_t$.
The unattenuated state $\ell_t$ is retained for subsequent
filter updates, including the computation of
$\gamma_{t+1}$.

\subsection{Tracking}
\label{sec:method:track}
For each frame, we extract ORB features using an eight-level image pyramid. The features support scale anchoring and projection-guided matching to local map points. These matches form the 2D–3D correspondences used for camera pose refinement.
Let $\boldsymbol{\mathcal{A}}_t$ denote the indices of these
correspondences.
For each matched landmark $\mathbf{X}_k$, its observed
pixel location is $\mathbf{u}_{tk}=(u_{tk},v_{tk})^\top$. We use the virtual stereo loss to optimize the camera pose as in ORB-SLAM3.
A valid corrected depth
$d_{tk}=\tilde{\mathbf{D}}_t(\mathbf{u}_{tk})$ defines a virtual
right-image coordinate
$
u_{R,tk}
=
u_{tk}-\frac{f_x b}{d_{tk}},
$
where $f_x$ is the horizontal focal length in $\mathbf{K}$.
The virtual baseline is initialized as
$b=0.0375\,d_{\mathrm{init}}$ and then held fixed,
where $d_{\mathrm{init}}$ is the median of the per-frame
median depths over the first eight frames.

Keeping the map points fixed, we refine the current pose
by minimizing a robust virtual stereo reprojection objective:

\begingroup
\begin{equation}
\begin{gathered}
\mathbf{T}_t =
\underset{\mathbf{T}\in SE(3)}{\arg\min}
\sum_{k\in\mathcal{A}_t}
\rho\!\left(
\frac{\|\mathbf{e}_{tk}(\mathbf{T})\|_2^2}
     {\eta_{tk}^2}
\right),
\\[0pt]
\mathbf{e}_{tk}(\mathbf{T}) =
\mathbf{x}_{tk}
-
\boldsymbol{\pi}_s(\mathbf{T}\mathbf{X}_k),
\end{gathered}
\label{eq:poseba}
\end{equation}
\endgroup
where $\mathbf{x}_{tk}=(u_{tk},v_{tk},u_{R,tk})^\top$ is the virtual
stereo observation, $\rho$ is the Huber robust loss, and
$\mathbf{T}\mathbf{X}_k
=\mathbf{R}^{\mathrm{cam}}\mathbf{X}_k+\mathbf{t}$
is the map point expressed in the camera frame of
$\mathbf{T}=(\mathbf{R}^{\mathrm{cam}},\mathbf{t})$.
The function
$\boldsymbol{\pi}_s(x,y,z)
=(f_x x/z+c_x,\ f_y y/z+c_y,\ f_x(x-b)/z+c_x)^\top$
is the virtual stereo projection with fixed intrinsics $\mathbf{K}$.
Following the virtual stereo observation model, each squared
residual norm is weighted by $\eta_{tk}^{-2}$, where
$\eta_{tk}$ is the pixel standard deviation associated with
the feature's pyramid level.

\textbf{Local bundle adjustment} jointly refines the pose
of a newly inserted tracking keyframe, the poses of
covisible keyframes, and their associated landmarks,
following ORB-SLAM3~\citep{campos2021orbslam3}.
Covisible keyframes are those that share observed landmarks.
Additional keyframes observing these local landmarks
contribute constraints while their poses remain fixed,
anchoring the local optimization to the surrounding map.
This joint refinement updates the sparse map
$\boldsymbol{\mathcal{M}}$ and improves consistency across keyframe
observations.
The updated geometry supports subsequent tracking and
depth-scale estimation, while revised camera poses are
propagated to the Gaussian mapper through the
synchronization step.

\textbf{Loop detection} retrieves candidate keyframes
using a bag-of-words representation of ORB descriptors. We adopt ORB-SLAM3's loop closure pipeline~\citep{campos2021orbslam3}
to reduce accumulated drift when revisiting previously
mapped regions. Loop correction aligns the revisited regions
and fuses duplicate landmarks.
The resulting corrections update keyframe poses
$\mathbf{T}_j$ and landmark positions $\mathbf{X}_k$
in the sparse map $\boldsymbol{\mathcal{M}}$.
Global bundle adjustment subsequently refines the
keyframe poses and landmarks in a background thread.
Before optimizing the next keyframe, the GS mapper updates
the poses of its stored keyframes with the corrected poses
from loop closure and applies the corresponding rigid
transformations to their associated Gaussians.

\subsection{Gaussian Mapper}
\label{sec:method:mapper}

The mapper incrementally updates the Gaussian map
$\boldsymbol{\mathcal{G}}$ defined in
Sec.~\ref{sec:method:representation}.
At each keyframe, render-aware Gaussian insertion
is followed by refinement with keyframe replay,
following OTF~\citep{meuleman2025onthefly}.
All updates use only the current observation and
previously stored keyframes.
Each stored keyframe contains its RGB image $\mathbf{I}_j$,
corrected depth $\tilde{\mathbf{D}}_j$, and camera pose
$\mathbf{T}_j$.  
OTF~\citep{meuleman2025onthefly} initializes poses using
mini bundle adjustment and subsequently refines them
jointly with Gaussians, without explicit loop closure.
Unlike OTF, in D4GS, historical keyframe poses are maintained by
the tracker through local bundle adjustment and loop
closure.
Before the Gaussian refinement, the mapper
synchronizes its stored keyframe poses with the latest
tracking poses.
For loop-closure corrections, it additionally applies
the corresponding rigid transformations to the Gaussians
associated with each keyframe, preserving their positions
and orientations relative to that keyframe.
This brings the Gaussian map into alignment with the
corrected cameras before further refinement.

\begin{figure*}[t]
    \centering
    \includegraphics[width=1\linewidth]{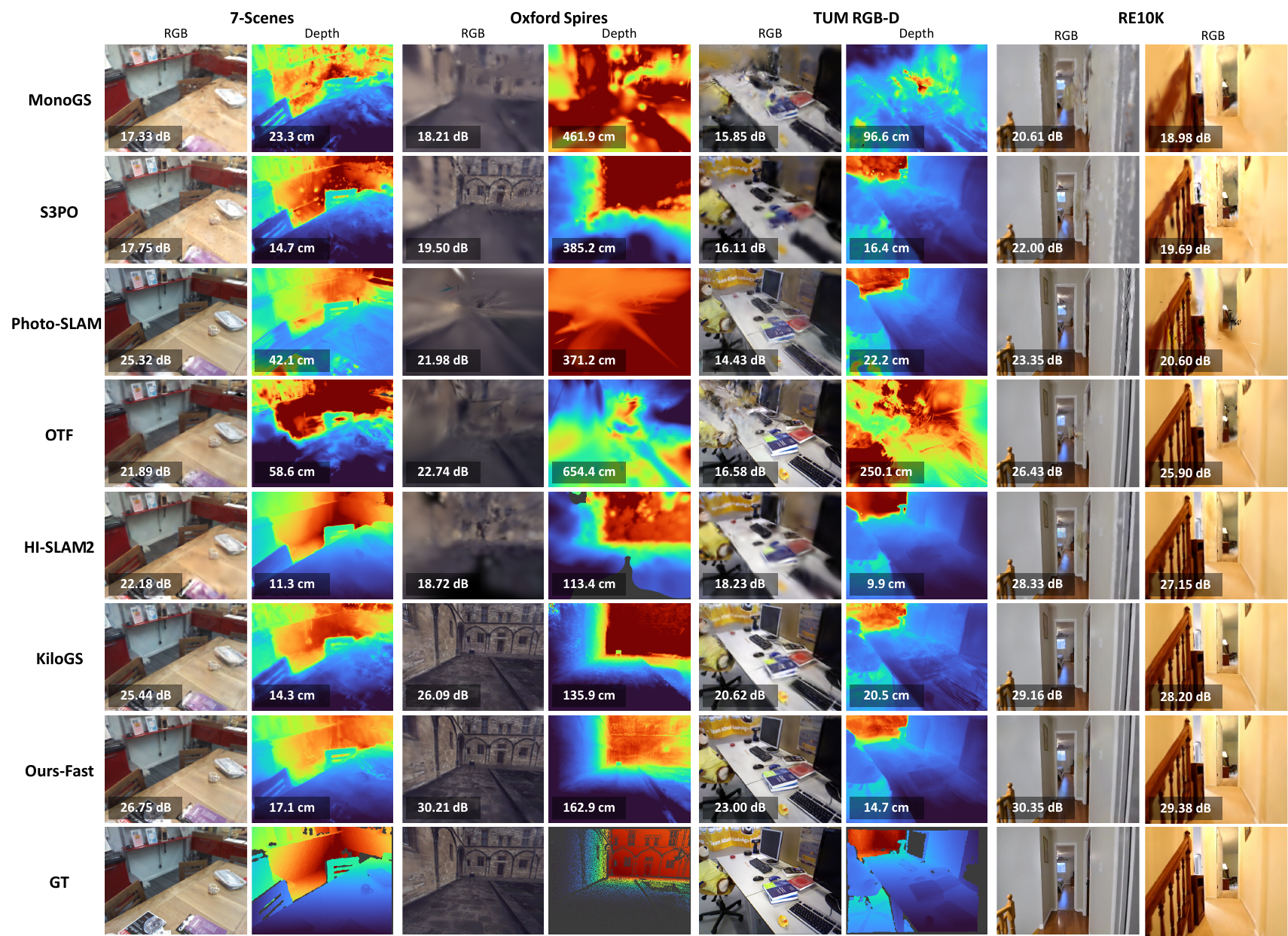}
    \caption{Qualitative rendering and depth comparison of Ours-Fast version.
One representative sequence is selected from each dataset.
For 7-Scenes, TUM-D, and Oxford Spires, we show both novel-view RGB rendering
and rendered depth; RE10K is shown only in RGB because reference depth
is unavailable. The 7-Scenes and TUM-D datasets provide dense reference depth,
whereas the Oxford Spires reference depth is sparsely projected from
the scanned point cloud. All methods are compared using the same
valid-depth mask.}
\label{fig:fast-vis}
\end{figure*}

\textbf{Laplacian-guided Gaussian insertion} allocates
new Gaussians preferentially to image structure that
is insufficiently represented by the current map.
Let $\hat{\mathbf{I}}$ denote the rendered RGB image at pose
$\mathbf{T}_t$, and let $\mathbf{u}=(u,v)^\top$
denote a pixel location.
We use $|\mathrm{Lap}(\mathbf{I})|$ to denote the absolute discrete
Laplacian of the grayscale version of $\mathbf{I}$, normalized
by its maximum, with a zero output when the maximum
is zero.
The insertion weights are
\begingroup
\begin{equation}
w(\mathbf{u})
\propto
\left[
|\mathrm{Lap}(\mathbf{I}_t)(\mathbf{u})|
-
|\mathrm{Lap}(\hat{\mathbf{I}})(\mathbf{u})|
\right]^+,
\label{eq:laplacian_insertion}
\end{equation}
\endgroup
where $[x]^+=\max\{0,x\}$.
Sampling is restricted to pixels with valid corrected
depth $\tilde{\mathbf{D}}_t(\mathbf{u})$.
This score favors pixels whose observed high-frequency
response exceeds that of the rendering, providing a
heuristic for identifying missing or under-resolved
detail.
For an empty map, $\hat{\mathbf{I}}$ is zero, and the weights
reduce to the observed image Laplacian over valid-depth
pixels.

\textbf{Geometric initialization} converts each sampled
pixel into a Gaussian. Since $\mathbf{T}_t$ maps world coordinates to camera
coordinates, the new Gaussian center is initialized by back-projection
$\boldsymbol{\mu}_i
=
\mathbf{T}_t^{-1}\cdot
\boldsymbol{\pi}^{-1}\!\left(\mathbf{u},\tilde{\mathbf{D}}_t(\mathbf{u})\right)$, where $\boldsymbol{\pi}^{-1}(\mathbf{u},z)=z\mathbf{K}^{-1}{\mathbf{u}}$.
Its remaining attributes are initialized as
$
\mathbf{q}_i = (1,0,0,0)^\top,\quad
\mathbf{s}_i = \frac{2z}{f_x}\mathbf{1}_3,\quad
\alpha_i = 0.07,\quad
\mathbf{c}_i = \mathbf{I}_t(\mathbf{u}).
$
The low initial opacity limits the contribution of
newly inserted Gaussians before refinement.

\textbf{Refinement with keyframe replay} updates
$\boldsymbol{\mathcal{G}}$ for 30 optimization steps after insertion.
At each step, we select the newest keyframe with
probability $p_{\mathrm{last}}=0.2$. Otherwise, we sample uniformly from all stored keyframes, including the newest.

For the sampled keyframe $j$, the differentiable
\texttt{gsplat} rasterizer~\citep{ye2025gsplat} produces
a color image $\hat{\mathbf{I}}_j$ and a rendered depth map
$\mathbf{z}_{\mathrm{render},j}$ at pose $\mathbf{T}_j$.
We minimize
\begingroup
\begin{equation}
\mathcal{L}_j =
0.8 L_{1,j}
+ 0.2 \bigl(1-\mathrm{SSIM}(\hat{\mathbf{I}}_j,\mathbf{I}_j)\bigr)
+ \lambda_d L_{\mathrm{depth},j},
\label{eq:mapping_loss}
\end{equation}
\endgroup
where $\lambda_d=0.01$ and
$L_{1,j}$ is the mean absolute RGB reconstruction error over the
image pixel. The depth loss compares the rendered depth with the stored
scale-corrected depth prior $\tilde{\mathbf{D}}_j$
in inverse-depth space:
\begingroup
\begin{equation}
L_{\mathrm{depth},j}
=
\frac{1}{|\boldsymbol{\mathcal{V}}_j|}
\sum_{\mathbf{u}\in\boldsymbol{\mathcal{V}}_j}
\left|
\frac{1}{\mathbf{z}_{\mathrm{render},j}(\mathbf{u})}
-
\frac{1}{\tilde{\mathbf{D}}_j(\mathbf{u})}
\right|,
\label{eq:inverse_depth_loss}
\end{equation}
\endgroup
where $\boldsymbol{\mathcal{V}}_j$ denotes the set of valid
depth pixels with positive, finite rendered
and reference depths.

\subsection{Edge Device Test and TensorRT Acceleration}
Test on Jetson Thor.

\section{Experiment}
\label{sec:exp}

\subsection{Setup and Evaluation}
\label{sec:exp:setup}

\paragraph{Datasets.}
We evaluate on four datasets, including indoor and outdoor datasets.
\textit{7-Scenes}~\citep{shotton2013scene} has seven indoor RGB
sequences with ground truth Kinect depth.
\textit{Oxford Spires}~\citep{tao2025spires} consists of six outdoor clips of
1000--2001 frames selected from outdoor sites: Keble College-(02,03,04), Observatory Quarter-01, Christ Church-05, and Bodleian Library-02, with poses and depth reprojected from the point cloud of
a terrestrial laser scan.
From \textit{RE10K}~\citep{zhou2018stereo} we use 16 short indoor and outdoor real-estate clips over 200 frames from the first four processed test chunks~\citep{charatan2024pixelsplat}. Five handheld sequences ( fr1\_desk, fr1\_desk2, fr2\_hemisphere, fr2\_desk, fr3\_long\_office) are selected from the \textit{TUM-D}~\citep{sturm2012benchmark} dataset, including the challenging loop inside the data. All frames are processed at 518 x 392 resolution across all datasets. We evaluate all methods using a single NVIDIA RTX 4090 GPU with an AMD Ryzen Threadripper PRO 3995WX 64-core CPU.

\textbf{Evaluation metrics} cover novel-view synthesis (NVS) quality, trajectory accuracy, reconstruction geometry, and end-to-end throughput.
We evaluate NVS quality on held-out views using PSNR, SSIM, and LPIPS.
We measure global trajectory accuracy using the root-mean-square absolute trajectory error (ATE) after aligning the estimated trajectory to the reference trajectory with a Sim(3) Umeyama alignment~\citep{umeyama1991least}.
For reconstruction geometry, we report the mean L1 error between rendered and ground-truth depth on held-out views after per-scene median scale alignment.
Finally, we measure end-to-end throughput (FPS) as the number of input frames divided by the total reconstruction time.

\textbf{Baselines} include MonoGS~\citep{matsuki2024monogs}, S3PO-GS~\citep{cheng2025s3po}, Photo-SLAM~\citep{huang2024photoslam}, OTF~\citep{meuleman2025onthefly}, HI-SLAM2~\citep{zhang2025hi}, KiloGS~\citep{yu2026robust}, DROID-Splat~\citep{homeyer2024droidsplat}, and ARTDECO~\citep{li2025artdeco}.
These baselines span different tracking and reconstruction strategies, including direct Gaussian-based tracking, feature-based tracking, and learned dense tracking coupled with Gaussian mapping.
They all emphasize SLAM or incremental reconstruction from unposed images.
All baselines are evaluated using their publicly available implementations on identical input frames and intrinsics.
For Photo-SLAM, we increase the mapping-iteration budget from 8 to at most 200 while retaining concurrent tracking and mapping. With this setting, it processes approximately 6 FPS. We evaluate two configurations of our system: Ours-Fast and Ours-Quality, which use 30 and 150 mapping iterations per keyframe, respectively.

\begin{table*}[t]
\centering
\definecolor{rkfirst}{RGB}{248,180,180}
\definecolor{rksecond}{RGB}{252,215,170}
\definecolor{rkthird}{RGB}{255,245,175}
\caption{Rendering quality, trajectory accuracy, geometry accuracy, and efficiency on all datasets. The \colorbox{rkfirst}{best}, \colorbox{rksecond}{second-best}, and \colorbox{rkthird}{third-best} results are highlighted. ATE and Depth L1 are in centimeters. Photo-SLAM fails on the Pumpkin sequence of 7-Scenes and is averaged over 6 scenes; OTF fails on one RE10K sequence and is averaged over 15 scenes. ARTDECO runs out of GPU memory (24\,GB) on Oxford Spires. RE10K does not have depth for evaluation.}
\label{tab:main}
\setlength{\tabcolsep}{3.6pt}
\resizebox{\textwidth}{!}{%
\begin{tabular}{l*{6}{c}@{\hspace{10pt}}*{6}{c}}
\toprule
\multirow{2}{*}{Method} & \multicolumn{6}{c}{\textbf{(a) 7-Scenes}} & \multicolumn{6}{c}{\textbf{(b) Oxford Spires}} \\
\cmidrule(lr){2-7}\cmidrule(lr){8-13}
 & PSNR$\uparrow$ & SSIM$\uparrow$ & LPIPS$\downarrow$ & ATE$\downarrow$ & Depth L1$\downarrow$ & FPS$\uparrow$ & PSNR$\uparrow$ & SSIM$\uparrow$ & LPIPS$\downarrow$ & ATE$\downarrow$ & Depth L1$\downarrow$ & FPS$\uparrow$ \\
\midrule
MonoGS & 18.94 & 0.681 & 0.501 & 29.90 & 51.75 & 3.28 & 19.63 & 0.608 & 0.586 & 872.56 & 989.76 & 3.05 \\
S3PO-GS & 19.76 & 0.713 & 0.470 & 7.30 & 16.84 & 1.74 & 21.29 & 0.689 & 0.512 & 426.46 & 484.01 & 1.26 \\
Photo-SLAM & 21.91 & 0.741 & 0.400 & 19.78 & 35.56 & \cellcolor{rkthird}5.99 & 21.35 & 0.690 & 0.496 & 749.13 & 981.92 & \cellcolor{rkthird}5.99 \\
OTF & 20.50 & 0.744 & 0.443 & 39.34 & 99.02 & \cellcolor{rksecond}7.57 & 22.72 & 0.701 & 0.476 & 954.29 & 1684.76 & \cellcolor{rksecond}8.24 \\
HI-SLAM2 & 21.50 & 0.739 & 0.461 & \cellcolor{rksecond}5.45 & \cellcolor{rkthird}15.49 & 1.32 & 21.09 & 0.656 & 0.551 & 284.91 & 374.42 & 0.95 \\
KiloGS & 23.48 & 0.784 & 0.362 & 12.92 & 16.63 & 1.07 & 25.30 & 0.818 & \cellcolor{rkfirst}0.237 & 202.11 & \cellcolor{rkthird}366.16 & 0.47 \\
DROID-Splat & \cellcolor{rkthird}25.78 & 0.853 & 0.302 & \cellcolor{rkfirst}5.17 & 19.14 & 1.61 & \cellcolor{rkfirst}28.16 & \cellcolor{rksecond}0.842 & 0.336 & \cellcolor{rksecond}52.01 & \cellcolor{rkfirst}215.40 & 1.05 \\
ARTDECO & \cellcolor{rksecond}26.60 & \cellcolor{rksecond}0.868 & \cellcolor{rkthird}0.279 & \cellcolor{rkthird}5.71 & \cellcolor{rkfirst}13.93 & 3.04 & \textcolor{red!75!black}{OOM} & \textcolor{red!75!black}{OOM} & \textcolor{red!75!black}{OOM} & \textcolor{red!75!black}{OOM} & \textcolor{red!75!black}{OOM} & \textcolor{red!75!black}{OOM} \\
\midrule
Ours-Fast & 25.11 & \cellcolor{rkthird}0.857 & \cellcolor{rksecond}0.275 & 8.22 & \cellcolor{rksecond}14.88 & \cellcolor{rkfirst}13.45 & \cellcolor{rkthird}26.63 & \cellcolor{rkthird}0.828 & \cellcolor{rkthird}0.325 & \cellcolor{rkthird}95.54 & 384.05 & \cellcolor{rkfirst}10.13 \\
Ours-Quality & \cellcolor{rkfirst}26.90 & \cellcolor{rkfirst}0.872 & \cellcolor{rkfirst}0.242 & 8.71 & 15.91 & 5.23 & \cellcolor{rksecond}27.66 & \cellcolor{rkfirst}0.852 & \cellcolor{rksecond}0.291 & \cellcolor{rkfirst}49.76 & \cellcolor{rksecond}296.28 & 3.27 \\
\midrule\midrule
\multirow{2}{*}{Method} & \multicolumn{6}{c}{\textbf{(c) RE10K}} & \multicolumn{6}{c}{\textbf{(d) TUM-D}} \\
\cmidrule(lr){2-7}\cmidrule(lr){8-13}
 & PSNR$\uparrow$ & SSIM$\uparrow$ & LPIPS$\downarrow$ & ATE$\downarrow$ & Depth L1$\downarrow$ & FPS$\uparrow$ & PSNR$\uparrow$ & SSIM$\uparrow$ & LPIPS$\downarrow$ & ATE$\downarrow$ & Depth L1$\downarrow$ & FPS$\uparrow$ \\
\midrule
MonoGS & 20.73 & 0.725 & 0.386 & 27.15 & -- & 1.61 & 16.48 & 0.593 & 0.568 & 104.14 & 124.13 & 2.22 \\
S3PO-GS & 21.67 & 0.767 & 0.333 & 4.16 & -- & 1.51 & 16.87 & 0.612 & 0.536 & 47.05 & 34.90 & 1.24 \\
Photo-SLAM & 21.96 & 0.745 & 0.336 & 1.97 & -- & \cellcolor{rkthird}5.97 & 13.81 & 0.493 & 0.582 & 58.88 & 86.78 & \cellcolor{rkthird}5.99 \\
OTF & 23.52 & 0.818 & 0.288 & 26.37 & -- & \cellcolor{rksecond}6.35 & 19.09 & 0.694 & 0.463 & 88.72 & 145.94 & \cellcolor{rksecond}6.54 \\
HI-SLAM2 & 25.20 & 0.836 & 0.271 & \cellcolor{rkfirst}1.09 & -- & 1.95 & 19.31 & 0.669 & 0.505 & \cellcolor{rksecond}2.87 & \cellcolor{rkfirst}19.12 & 0.92 \\
KiloGS & 26.33 & 0.869 & \cellcolor{rksecond}0.170 & 4.69 & -- & 0.60 & 21.49 & 0.736 & 0.401 & 17.67 & \cellcolor{rkthird}22.02 & 0.77 \\
DROID-Splat & \cellcolor{rkthird}28.11 & \cellcolor{rksecond}0.906 & 0.180 & \cellcolor{rksecond}1.43 & -- & 1.98 & 22.96 & 0.807 & 0.336 & \cellcolor{rkfirst}2.69 & 25.99 & 1.06 \\
ARTDECO & \cellcolor{rkfirst}28.34 & \cellcolor{rkthird}0.902 & \cellcolor{rkthird}0.176 & \cellcolor{rkthird}1.56 & -- & 2.28 & \cellcolor{rksecond}24.59 & \cellcolor{rksecond}0.837 & \cellcolor{rkthird}0.303 & \cellcolor{rkthird}3.85 & \cellcolor{rksecond}21.59 & 2.07 \\
\midrule
Ours-Fast & 26.00 & 0.879 & 0.195 & 2.43 & -- & \cellcolor{rkfirst}10.05 & \cellcolor{rkthird}23.36 & \cellcolor{rkthird}0.828 & \cellcolor{rksecond}0.283 & 5.97 & 26.65 & \cellcolor{rkfirst}11.29 \\
Ours-Quality & \cellcolor{rksecond}28.33 & \cellcolor{rkfirst}0.909 & \cellcolor{rkfirst}0.143 & 2.52 & -- & 4.68 & \cellcolor{rkfirst}24.73 & \cellcolor{rkfirst}0.849 & \cellcolor{rkfirst}0.250 & 5.38 & 25.21 & 3.93 \\
\bottomrule
\end{tabular}}
\end{table*}

\begin{figure*}[t]
    \centering
    \includegraphics[width=1\linewidth]{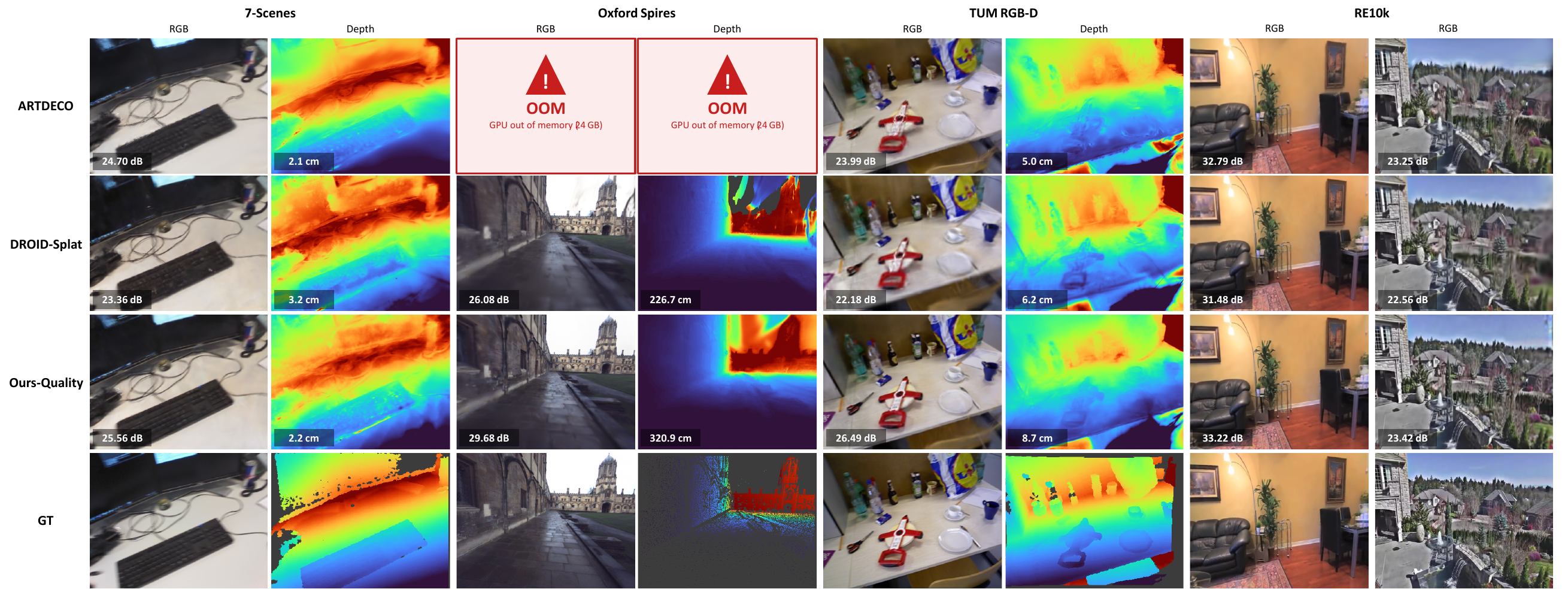}
    \caption{Qualitative rendering and depth comparison of Ours-Quality version.
One representative sequence is selected from each dataset.
For 7-Scenes, TUM-D, and Oxford Spires, we show both novel-view RGB rendering
and rendered depth; RE10K is shown only in RGB because reference depth
is unavailable. ARTDECO fails on the Oxford Spires due to OOM.}
\label{fig:quality-vis}
\end{figure*}

\subsection{Quantitative and Qualitative Results}

\textbf{Ours-Fast} achieves strong novel-view synthesis quality
while providing the highest throughput among all evaluated
methods (Tab.~\ref{tab:main}).
With only 30 mapping iterations per keyframe, it runs at
10.05--13.45\,FPS and generally outperforms MonoGS, S3PO-GS,
Photo-SLAM, OTF, HI-SLAM2, and KiloGS in rendering quality.
In particular, it achieves higher PSNR and SSIM and lower
LPIPS than the first five methods on all four datasets.
Compared with KiloGS, it achieves higher SSIM throughout,
although KiloGS retains a small advantage in LPIPS on Oxford
Spires and in PSNR and LPIPS on RE10K.
Fig.~\ref{fig:fast-vis} illustrates this advantage:
Ours-Fast preserves sharper object boundaries and finer
texture details, with fewer visible rendering artifacts
in the novel views.
These results demonstrate that a small mapping budget
is sufficient for high-quality rendering at substantially
higher throughput across diverse scenes.

\textbf{Ours-Quality} further improves rendering quality
to compete with DROID-Splat and ARTDECO, which achieve
strong tracking accuracy but operate at approximately
1--3\,FPS.
By increasing the mapping budget to 150 iterations per
keyframe only, Ours-Quality
achieves the best SSIM on all four datasets and the best
PSNR and LPIPS on 7-Scenes and TUM-D.
As illustrated in Fig.~\ref{fig:quality-vis}, its renderings
preserve visual detail competitive with or better than
them.
Its depth reconstruction remains competitive: Depth~L1
is lower than that of DROID-Splat on 7-Scenes and TUM-D,
although higher on Oxford Spires and higher than that
of ARTDECO on the two indoor datasets.
For ATE, Ours-Quality achieves the lowest
ATE on Oxford Spires, while DROID-Splat and ARTDECO
retain an advantage on the remaining datasets. Notably, ARTDECO fails on the Oxford Spires due to out-of-memory errors.
Despite these trade-offs, Ours-Quality runs at
3.27--5.23 FPS, exceeding both methods wherever they
complete reconstruction.
Together, the two configurations provide a flexible
trade-off between rendering quality and computational
cost within a single pipeline.

\begin{table}[t]
\centering
\caption{Runtime on 7-Scenes with 30 and 150 mapping iterations. All times in ms.
$^\dagger$Runs asynchronously, overlapped with depth inference
of the next frame.}
\label{tab:timing}

\renewcommand{\arraystretch}{0.9}\footnotesize
\setlength{\aboverulesep}{1pt}
\setlength{\belowrulesep}{1pt}

\begin{tabular*}{\columnwidth}{@{\extracolsep{\fill}}lrr@{}}
\toprule
 & Ours-Fast & Ours-Quality \\
\midrule
Depth inference       & 29.6  & 29.4 \\
Tracking$^\dagger$     & 24.8  & 25.0 \\
Frame latency         & 32.7  & 32.5 \\
Mapping per keyframe  & 417   & 1226 \\
\midrule
End-to-end FPS        & 13.45 & 5.23 \\
\bottomrule
\end{tabular*}
\end{table}

\subsection{Timing Breakdown}
\label{sec:timing}

Tab.~\ref{tab:timing} breaks down the runtime on 7-Scenes. The front end
is real-time: depth inference takes 29.6\,ms per frame, and ORB-SLAM3
tracking with scale alignment runs asynchronously, overlapped with depth
inference of the next frame, giving a frame latency of 32.7\,ms
($\sim$30\,FPS). Tracking and mapping run in two parallel threads that share one GPU.  Mapping is the bottleneck: it runs only on keyframes
($\sim$15\% of the frames), but each one costs 417\,ms at 30 iterations,
i.e.\ 62.7\,ms per frame after amortization. Since the mapping cost grows
linearly with the iteration budget while the tracking cost stays constant,
the iteration budget is the single knob that trades speed
(13.45 $\rightarrow$ 5.23\,FPS) for rendering quality, with no change to the rest of the pipeline.

\begin{table}[!htb]
\centering
\caption{Ablation on monocular depth source over 7-Scenes.
Params in M, time in ms, ATE in cm.}
\label{tab:abl_depth}
\footnotesize
\setlength{\tabcolsep}{0pt}
\begin{tabular*}{\columnwidth}{@{\extracolsep{\fill}}lcccccc@{}}
\toprule
Depth source & Params$\downarrow$ & Time$\downarrow$ & PSNR$\uparrow$ & SSIM$\uparrow$ & LPIPS$\downarrow$ & ATE$\downarrow$ \\
\midrule
Metric3DV2   & 1377.7 & 191.6 & 25.15 & 0.857 & 0.275 & 7.81 \\
LingBot-Map  & 1157.9 & 131.0 & \textbf{25.33} & \textbf{0.862} & \textbf{0.266} & \textbf{6.58} \\
UniDepthV2   & 353.8  & 45.8  & 25.22 & 0.861 & 0.269 & 7.20 \\
DA2-Metric   & 335.3  & 48.2  & 25.10 & 0.857 & 0.277 & 7.67 \\
\midrule
DA3-Metric-L & \textbf{334.2} & \textbf{29.6} & 25.11 & 0.857 & 0.275 & 8.22 \\
\bottomrule
\end{tabular*}
\end{table}

\subsection{Ablation Study}

We conduct extensive ablation studies on scale anchoring,
depth sources, pose solvers, and loop closure to evaluate
their respective contributions to system performance.

\begin{table}[!htb]
\centering
\caption{Ablation of the online scale anchoring on 7-Scenes.}
\label{tab:abl_kpmt}
\setlength{\tabcolsep}{4pt}
\resizebox{\columnwidth}{!}{%
\begin{tabular}{@{}lcccc@{}}
\toprule
Scale anchoring
& PSNR$\uparrow$
& SSIM$\uparrow$
& LPIPS$\downarrow$
& ATE (cm)$\downarrow$ \\
\midrule
w/o SA
& 24.85 & 0.848 & 0.288 & 8.81 \\
Ours
& \textbf{25.11}
& \textbf{0.857}
& \textbf{0.275}
& \textbf{8.22} \\
\bottomrule
\end{tabular}%
}
\end{table}

\textbf{Ablation on scale anchoring} in
Tab.~\ref{tab:abl_kpmt} isolates the contribution of scale
anchoring by comparing scale-aligned depth with raw predicted
depth while keeping the remaining components unchanged.
Enabling scale anchoring improves all four metrics on
7-Scenes, highlighting the importance of aligning predicted
depth with the scale of the existing map.
These results suggest that reducing scale discrepancies
benefits both camera tracking and Gaussian reconstruction.

\textbf{Ablation on depth source} in Tab.~\ref{tab:abl_depth}
replaces the depth predictor while keeping the rest of the system fixed. Rendering quality is largely insensitive to the depth source:
PSNR spans only 0.23\,dB (25.10--25.33) across five networks
whose size differs by $4\times$ and whose speed differs by $6\times$.
We therefore use DA3-Metric-L by default.
It is the smallest (334\,M parameters) and fastest (29.6\,ms)
model. The larger LingBot-Map gives the lowest ATE, but at
$3.5\times$ the parameters and $4\times$ the per-frame time.

\begin{table}[!htb]
\centering
\caption{Ablation on different camera tracking over 7-Scenes. For variant C, we use Open3D's point-to-plane ICP implementation.}
\label{tab:7s_tracking}
\setlength{\tabcolsep}{3pt}
\renewcommand{\arraystretch}{1.1}
\resizebox{\columnwidth}{!}{%
\begin{tabular}{@{}lccccc@{}}
\toprule
Tracker
& PSNR$\uparrow$
& SSIM$\uparrow$
& LPIPS$\downarrow$
& ATE (cm)$\downarrow$
& Track (ms)$\downarrow$ \\
\midrule
A: Ours
& \textbf{25.11}
& \textbf{0.857}
& \textbf{0.275}
& \textbf{8.22}
& \underline{24.8} \\

B: 2D reprojection
& \underline{22.13}
& \underline{0.784}
& \underline{0.368}
& \underline{11.69}
& \textbf{21.8} \\

C: ICP solver
& 18.36
& 0.660
& 0.522
& 25.27
& 94.0 \\
\bottomrule
\end{tabular}%
}
\end{table}

\textbf{Ablation on pose solver} compares our virtual stereo
formulation (A), 2D reprojection using depth only for map-point
initialization and omitting the depth constraint in
Eq.~\ref{eq:poseba} (B), and point-to-plane ICP replacing
pose optimization within the same ORB framework (C).
For C, we adopt the CPU implementation of point-to-plane ICP
provided by Open3D, without additional runtime optimization.
As shown in Table~\ref{tab:7s_tracking}, A achieves the best
rendering quality and lowest ATE with only 3\,ms of overhead
relative to B, while C is less accurate and takes longer.

\begin{figure}[!htb]
    \centering
    \includegraphics[width=\columnwidth]{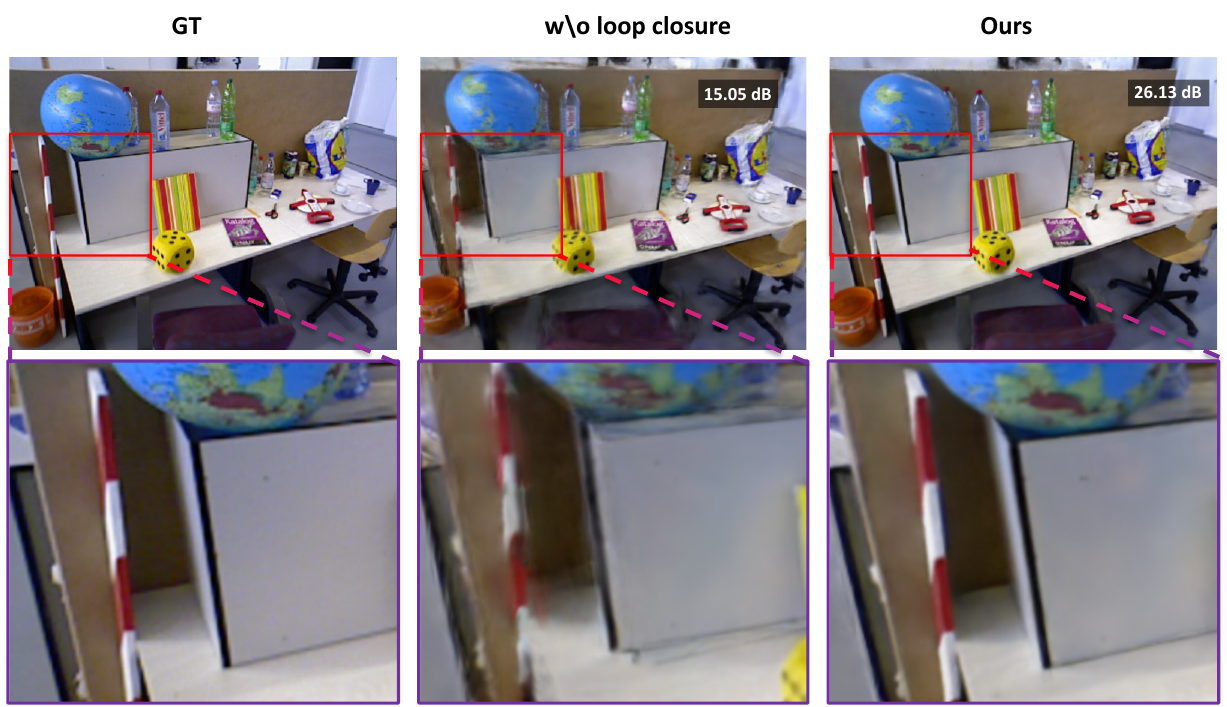}
    \caption{Qualitative comparison with and without loop closure
    on fr3\_long\_office. Loop closure reduces ghosting and floating
    artifacts caused by misaligned surfaces in revisited regions.}
    \label{fig:loop_qual}
\end{figure}

\textbf{Ablation on loop closure} in Tab.~\ref{tab:abl_loop}
evaluates its effect on TUM RGB-D.
Averaged over the five sequences, enabling loop closure improves
PSNR by 0.41\,dB
and reduces ATE from 6.62 to 5.97\,cm ($-$9.8\%).
As shown in Fig.~\ref{fig:loop_qual}, loop closure reduces
ghosting and floating artifacts caused by misaligned surfaces.

\begin{table}[!htb]
\centering
\caption{Ablation of loop closure on TUM RGB-D.}
\label{tab:abl_loop}
\small    
\setlength{\tabcolsep}{3pt}
\renewcommand{\arraystretch}{1.1}
\resizebox{\columnwidth}{!}{%
\begin{tabular}{lcccc}
\toprule
Method & PSNR$\uparrow$ & SSIM$\uparrow$ & LPIPS$\downarrow$ & ATE (cm)$\downarrow$ \\
\midrule
w/o loop closure & 22.95 & 0.819 & 0.293 & 6.62 \\
Ours             & \textbf{23.36} & \textbf{0.828} & \textbf{0.283} & \textbf{5.97} \\
\bottomrule
\end{tabular}
}
\end{table}

\section{Conclusion}
\label{sec:con}

We present D4GS, an online monocular Gaussian SLAM system
that combines learned geometry prediction with sparse
feature-based camera tracking and incremental Gaussian mapping.
By aligning predicted depth with a sparse map and filtering
the scale correction over time, D4GS provides a shared depth
prior for tracking and mapping.
Experiments on four datasets spanning diverse indoor and
outdoor scenes demonstrate an improved trade-off between
rendering quality and throughput over previous methods.
The diversity of the evaluated environments supports the
applicability of D4GS across a range of scenarios,
highlighting its potential for practical reconstruction tasks.
Our ablation with five depth predictors further demonstrates
the framework's compatibility with different sources of
learned geometry.

The current system can not handle dynamic scenes, and its tracking performance degrades in environments with repetitive or weak textures. Addressing these limitations while preserving efficient online operation is an important direction for future work.

\section{LLM Usage Statement}
In this work, we used generative AI tools (ChatGPT and Claude Code) to provide feedback on research methodology and experiments and to assist with the implementation of methods. Additionally, we used generative AI tools for brainstorming, code editing, literature search, and improving the readability of the manuscript. All AI-assisted methodological suggestions, code, experimental results, and manuscript content were independently reviewed and verified by the authors. We take responsibility for the final content of this work.
\section*{Acknowledgements}
This work was supported by the Guangdong Basic and Applied Basic Research Foundation (No. 2026A1515011138) and the King Abdullah University of Science and Technology (KAUST)  research fund under grant BAS/1/1680-01-01.
\printbibliography

\clearpage
\appendix

\section{Scale anchoring Details}
\label{supp:scale_anchoring}

Our scale anchoring module estimates a frame-wise depth
correction relative to the sparse map and regularizes it
over time. The corrected depth is shared by camera
tracking and Gaussian mapping.

\textbf{Scale observation.}
We predict the current world-to-camera pose as
$\bar{\mathbf{T}}_t
=\mathbf{V}_{t-1}\mathbf{T}_{t-1}$,
where
$\mathbf{V}_{t-1}
=\mathbf{T}_{t-1}\mathbf{T}_{t-2}^{-1}$
is the relative motion between the two most recent
estimated poses, assumed constant for the next frame.
Previous inlier landmarks are projected into the current
image and matched to nearby ORB features using descriptors,
without depth-based consistency gating.

For each matched landmark $\mathbf{X}_k$, we compare
its predicted camera-space depth with the network depth:
\begin{equation}
\bar{z}_{tk}
=
[\bar{\mathbf{R}}_t^{\mathrm{cam}}\mathbf{X}_k
+\bar{\mathbf{t}}_t]_z,
\qquad
r_{tk}
=
\log\frac{\bar{z}_{tk}}
{\mathbf{D}_t(\mathbf{u}_{tk})}.
\label{eq:supp_scale_ratio}
\end{equation}
Only finite, positive depths are retained.
Matches with descriptor Hamming distance
$h_{tk}>100$ are rejected.
The scale observation $m_t$ is the weighted median
of the valid log-depth ratios, with weights
\begin{equation}
w_{tk}
=
\max\left\{10^{-3},\,1-\frac{h_{tk}}{100}\right\}.
\label{eq:supp_scale_weight}
\end{equation}
Specifically, we sort the ratios and select the first
value whose cumulative weight reaches half of the total
weight; explicit weight normalization is unnecessary.

\textbf{Temporal filtering.}
We maintain a scalar log-scale state $\ell_t$
and its variance $P_t$.
Observation reliability is estimated using the
unweighted median absolute deviation (MAD):
\begin{equation}
\begin{gathered}
r_t^{\mathrm{med}}
=\operatorname{median}_k r_{tk},\\
\delta_t
=\operatorname{median}_k
  |r_{tk}-r_t^{\mathrm{med}}|,\\
\nu_t
=\max\{10^{-6},(1.4826\,\delta_t)^2\}.
\end{gathered}
\label{eq:supp_scale_mad}
\end{equation}
The factor $1.4826$ converts MAD into a
standard-deviation estimate under a Gaussian assumption.
We use this dispersion as an observation-variance proxy
without dividing by the match count, since errors
within a frame can be correlated.

Assuming a random walk with process variance $q^2$,
the Kalman update is
\begin{equation}
\begin{gathered}
P_t^- = P_{t-1}+q^2,
\qquad
\kappa_t = \frac{P_t^-}{P_t^-+\nu_t},\\
\ell_t = \ell_{t-1}
+\kappa_t(m_t-\ell_{t-1}),\\
P_t = (1-\kappa_t)P_t^-.
\end{gathered}
\label{eq:supp_scale_filter}
\end{equation}
If motion prediction is unavailable or fewer than
40 valid matches remain, we skip the measurement update:
$\ell_t=\ell_{t-1}$ and
$P_t=P_t^-=P_{t-1}+q^2$.

\textbf{Scale attenuation and application.}
New landmarks inherit the corrected depth, so repeated
alignment to the evolving sparse map can accumulate
scale drift.
We therefore use the approximately metric network
prediction as a soft reference by attenuating large
log-scale corrections:
\begin{equation}
\gamma_t
=
\frac{\sigma^2}{\sigma^2+\ell_{t-1}^2},
\qquad
\tilde{\mathbf{D}}_t
=
\exp(\gamma_t\ell_t)\mathbf{D}_t,
\qquad
\sigma=0.15.
\label{eq:supp_scale_application}
\end{equation}
Both tracking and mapping receive
$\tilde{\mathbf{D}}_t$, while the unattenuated state
$\ell_t$ is retained for subsequent filter updates
and the computation of $\gamma_{t+1}$.
This correction adjusts the global depth scale of each
frame; it does not correct spatially varying depth errors.

\section{Dataset Details}
We evaluate on four public datasets that are deliberately complementary:
room-scale indoor scenes with dense ground truth (7-Scenes), handheld
sequences that revisit previously mapped space (TUM RGB-D), uncontrolled
real-world video (RealEstate10K), and large-scale outdoor walks with
survey-grade ground truth (Oxford Spires).  This section lists every sequence
and frame range we use.  The same frames, held-out views, and camera intrinsics are given to every method.

\paragraph{7-Scenes~\citep{shotton2013scene}.}
We use all seven scenes (chess, fire, heads, office, pumpkin, redkitchen,
stairs) and, for each scene, the complete first sequence \texttt{seq-01}
(1000 frames; 500 for stairs) without subsampling.

\paragraph{TUM RGB-D~\citep{sturm2012benchmark}.}
We focus on long-term consistency, i.e.\ sequences in which the camera
returns to previously observed regions so that drift and loop closure can
be evaluated.  We use five such sequences, each in full:
\texttt{fr1\_desk} (596 frames), \texttt{fr1\_desk2} (640 frames),
\texttt{fr2\_360\_hemisphere} (2729 frames), \texttt{fr2\_desk} (2965 frames) and
\texttt{fr3\_long\_office\_household} (2488 frames).  The two \texttt{fr1}
sequences revisit a desk repeatedly under fast handheld motion; the other
three close a large loop over 16--22\,m.

\paragraph{Oxford Spires~\citep{tao2025spires}.}
We use the forward camera (\texttt{cam0}, 20\,Hz) of six recordings from
four sites.  The recordings are captured on public grounds with passers-by,
whereas all compared methods assume a static scene and the TLS ground truth
contains only static structure; we therefore take from each recording one
contiguous clip without moving objects, using all frames of the clip.  The
frame ranges $[start,end)$ in the \texttt{cam0} stream and the resulting numbers
of frames are:
\texttt{bodleian-library-02} [3689,\,4721) (1032 frames),
\texttt{christ-church-05} [647,\,1700) (1053 frames),
\texttt{keble-college-02} [0,\,2001) (2001 frames),
\texttt{keble-college-03} [2500,\,3500) (1000 frames),
\texttt{keble-college-04} [0,\,1000) (1000 frames) and
\texttt{observatory-quarter-01} [1208,\,2209) (1001 frames).

\paragraph{RealEstate10K~\citep{zhou2018stereo}.}
We use the test split as released by pixelSplat~\citep{charatan2024pixelsplat} (chunks
\texttt{000000}--\texttt{000004}).  Since most clips are short (median
${\approx}120$ frames), we keep the 16 clips with more than 200 frames
(212--279 frames), each in full, so that the sequences are long enough to
evaluate SLAM.  Clips are identified by chunk:index---\texttt{000000}: 4, 13,
15; \texttt{000001}: 3, 4, 7, 10; \texttt{000002}: 2, 3, 11, 12;
\texttt{000003}: 17; \texttt{000004}: 0, 2, 8, 9---corresponding to the
RE10K keys
\texttt{6558c5f10d45a929}, \texttt{20d86cff490c0c42}, \texttt{fea544b472e9abd1},
\texttt{c48f19e2ffa52523}, \texttt{1214f2a11a9fc1ed}, \texttt{21e794f71e31becb},
\texttt{6771a51bf0cfce7f}, \texttt{ffa95c3b40609c76}, \texttt{a47b88040452d7d9},
\texttt{8adebbb68f2c3f84}, \texttt{29e0bfbad00f0d5e}, \texttt{3087828bc27bc4c7},
\texttt{e124dfa7e996a414}, \texttt{79ce50fb7b021e1d}, \texttt{1f0e06e4388dd600}
and \texttt{c9c30723d7c54793}.

\section{Edge device Test}
\label{sec:edge}

Results obtained on desktop-class GPUs do not necessarily carry over to the
platforms on which a SLAM system is ultimately expected to run. Robotic
and autonomous platforms often require onboard computation under constrained
power and thermal budgets, making real-time performance on embedded hardware
important for practical deployment. To evaluate the deployment feasibility
of our system, we implement it on the NVIDIA Jetson AGX Thor, a high-end
embedded computing platform designed for robotics and physical AI.
It integrates a 14-core Arm Neoverse-V3AE CPU, a Blackwell-architecture
GPU, and 128\,GB of unified LPDDR5X memory, with a configurable power
range of 40--130\,W. This provides a representative high-end embedded
testbed for assessing the real-time capabilities of our system beyond
desktop-class GPUs.

Profiling the direct deployment shows that the per-frame depth network is the
dominant cost: a single depth inference takes about 60\,ms, which largely limits
the achievable frame rate. We therefore mainly accelerate this stage with CUDA Graphs,
which remove the per-kernel launch overhead, and TensorRT, which compiles the
network into an optimized inference engine. Together they reduce the depth
inference time from about 60\, ms to about 25\, ms. As reported in
Table~\ref{tab:edge_fps}, this raises the end-to-end frame rate, averaged over
the seven scenes of 7-Scenes, from 8.42 to 15.64\, FPS, a $1.86\times$ speedup.

\begin{table}[t]
  \centering
  \caption{End-to-end frame rate on the NVIDIA Jetson AGX Thor, averaged over
  the seven scenes of 7-Scenes.}
  \label{tab:edge_fps}
  \begin{tabular}{lc}
    \toprule
    Configuration & FPS \\
    \midrule
    Without acceleration & 8.42 \\
    With CUDA Graph + TensorRT & 15.64 \\
    \bottomrule
  \end{tabular}
\end{table}

\end{document}